\documentclass[letterpaper,10pt,conference]{ieeeconf}

\IEEEoverridecommandlockouts  
\usepackage{tikz}
\usetikzlibrary{shapes,arrows,fit,calc,positioning,automata,backgrounds,calc,patterns,decorations.pathmorphing,decorations.markings}
\usepackage[thinc]{esdiff}
\graphicspath{{figs/}}
\usepackage{graphicx, amssymb, array, rotating}
\usepackage{subfigure}
\usepackage{subcaption}
\usepackage[utf8]{inputenc}
\usepackage{svg}
\usepackage{siunitx}
\usepackage{booktabs}
\usepackage{xfrac}
\usepackage[draft, markup=bfit, deletedmarkup=sout, authormarkup=brackets]{changes}
\usepackage[thinc]{esdiff}
\usepackage{times}
\usepackage{multicol}
\usepackage[bookmarks=true]{}
\usepackage{xcolor, soul}
\usepackage{color}
\usepackage{graphics}
\usepackage{epsfig} 
\usepackage{caption}
\usepackage{mdwmath}
\usepackage{mdwtab}
\usepackage{url}
\usepackage{float}
\usepackage[makeroom]{cancel}
\usepackage{amsmath,bm}
\usepackage{dblfloatfix}
\usepackage{lipsum}
\usepackage{relsize}
\usepackage{algpseudocode}
\usepackage[ruled,vlined]{algorithm2e}
\usepackage{xspace}
\usepackage{mathtools}
\usepackage{amsfonts}
\usepackage[noadjust]{cite}
\usepackage{flushend}
\usepackage[normalem]{ulem} 
\usepackage{marginnote}
\usepackage{atbegshi}
\usepackage[font={small}]{caption}
\usepackage[draft, markup=bfit, deletedmarkup=sout, authormarkup=brackets]{changes}
\AtBeginDocument{\AtBeginShipoutNext{\AtBeginShipoutDiscard}}

\makeatletter
\let\NAT@parse\undefined
\makeatother

\usepackage[colorlinks=true,linkcolor=blue,urlcolor=blue,citecolor=blue,anchorcolor=blue]{hyperref} 
\renewcommand{\baselinestretch}{0.921}

\title{\LARGE \bf Visuo-Tactile Exploration of Unknown Rigid 3D Curvatures by Vision-Augmented Unified Force-Impedance Control}

\author{Kübra Karacan, Anran Zhang, Hamid Sadeghian, Fan Wu,  and Sami Haddadin}

\makeatletter
\def\BState{\State\hskip-\ALG@thistlm}
\makeatother      
   
\begin{document}
\thanks{The authors are with the Chair of Robotics and Systems Intelligence, MIRMI-Munich Institute of Robotics and Machine Intelligence, Technical University of Munich, Germany, and the Centre for Tactile Internet with Human-in-the-Loop (CeTI). 
{\tt\small kuebra.karacan@tum.de}}
\maketitle
\thispagestyle{empty}\pagestyle{empty}

\begin{abstract}
Despite recent advancements in torque-controlled tactile robots, integrating them into manufacturing settings remains challenging, particularly in complex environments. Simplifying robotic skill programming for non-experts is crucial for increasing robot deployment in manufacturing. This work proposes an innovative approach, Vision-Augmented Unified Force-Impedance Control (VA-UFIC), aimed at intuitive visuo-tactile exploration of unknown 3D curvatures. VA-UFIC stands out by seamlessly integrating vision and tactile data, enabling the exploration of diverse contact shapes in three dimensions, including point contacts, flat contacts with concave and convex curvatures, and scenarios involving contact loss. A pivotal component of our method is a robust online contact alignment monitoring system that considers tactile error, local surface curvature, and orientation, facilitating adaptive adjustments of robot stiffness and force regulation during exploration. We introduce virtual energy tanks within the control framework to ensure safety and stability, effectively addressing inherent safety concerns in visuo-tactile exploration. Evaluation using a Franka Emika research robot demonstrates the efficacy of VA-UFIC in exploring unknown 3D curvatures while adhering to arbitrarily defined force-motion policies. By seamlessly integrating vision and tactile sensing, VA-UFIC offers a promising avenue for intuitive exploration of complex environments, with potential applications spanning manufacturing, inspection, and beyond.
\end{abstract}


\section{Introduction}\label{sec:intro}
Robotic systems have become indispensable in industrial operations, excelling in tasks demanding repetitive speed and precision. However, challenges persist when these systems confront tasks requiring nuanced force and compliance control, such as polishing car doors or carving metal. Despite advancements in torque-controlled tactile robots, their deployment for tactile and flexible interaction remains limited due to the expertise required in control implementation~\cite{Billard2019}.

To enhance the deployment of tactile robots, the development of straightforward and intuitive robot skill programming methods is essential to alleviate the need for intricate tailoring and adjustment of software programs according to the task specifications of each application. In traditional factory settings, industry experts experienced in standard automation processes program the machines, such as CNC machines, to perform required motion or force to deliver high-quality operations~\cite{Lambert2004}. However, although robotics has made vast progress in force-motion interaction, including impedance, force, and unified controls~\cite{Hogan1984, khatib1995, Karacan2022}, in flexible manufacturing where frequent reconfiguration is common, it remains difficult to efficiently program the robots while adhering to desired forces and motions derived from task and process requirements. Moreover, deploying robots in highly variable environments, such as small batch-size production, requires fine-tuning robot controllers to adapt to changing environmental features and constraints~\cite{schutter2007, Kramberger2016, Iskandar2023}.
\begin{figure}
    \centering
      \includegraphics[width=0.95\linewidth]{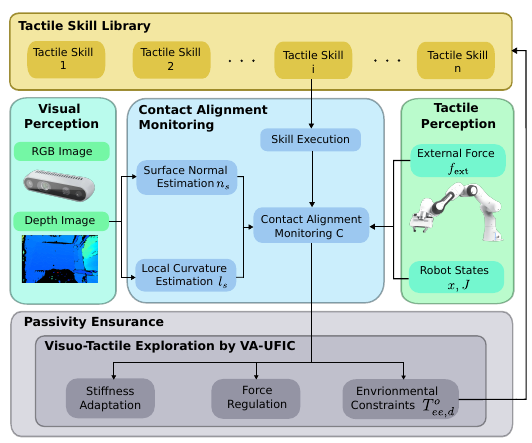}
          \vspace{-0.1cm} 
    \caption{ \textbf{Visuo-Tactile Exploration of Unknown Rigid 3D Curvatures by vision-augmented unified force-impedance control (VA-UFIC) for a chosen tactile skill.} Visuo-tactile exploration is the next step to achieving a force-motion planning framework that outputs an object-centric force-motion profile for an arbitrary tactile skill policy. The explored environment is fed back to the library to further plan the force-motion policy.}
    \label{fig:workflow}
\end{figure}

To achieve more natural and intuitive robot programming to broaden robot deployment in manufacturing, it is desirable to autonomously explore environmental features for a given arbitrary force-motion policy and use the explored environment information to plan the object-centric force-motion policy, as shown in Fig.~\ref{fig:workflow}. Methods such as the operational space framework, constrained-based task specifications, and object-centric representations constitute significant steps towards a user-friendly programming paradigm~\cite{Khatib1987, Balta2017, migimatsu2020, vochten2023}. However, directly producing or planning the object-centric force-motion policy for a non-control expert, given an arbitrary force-motion policy, requires autonomous investigation of the environmental constraints experienced by the tools, such as surface curvatures or normal, during task execution. This approach would allow non-experts to use controllers, leveraging environment exploration and analysis of current surface constraints.

Integrating visual and tactile sensors for contact alignment monitoring, like an intelligent end-effector, offers a promising solution to enhance robots' environmental awareness, particularly in exploring unknown surface constraints such as curvatures. While visual perception enables robots to perceive environmental details without touching, tactile sensors provide unique insights into force and moments not discernible through vision alone~\cite{Chi2018, Ganguly2020}. However, challenges arise when irregularities occur outside the camera's field of view, i.e., the camera's view is blocked in the contact point or when tactile sensors fail to sense forces and moments due to the point contact or even loss of contact. In other words, different contact shapes dictate the sensing modality for perceiving environmental features. Thus, unifying visual and tactile sensors to monitor the contact alignment between the tool and surface presents a more comprehensive solution involving various contact shapes in real-world applications. Approaches in robotics that synergise visual perception and tactile sensing vary, focusing on enhancing grasp stability, evaluating object shapes, or executing manipulation tasks based on predefined structures such as manipulation graphs or computer-aided design models~\cite{Calandra2018, Sachtler2019, Fazeli2019, Nottensteiner2021, Suresh2023}. Despite advancements in visuo-tactile capabilities, using those methods in environment exploration is mainly limited in 2D for specific contact shapes, persisting in a gap between current robotic capabilities and real-world application demands~\cite{Suresh2021, Kato2022, Qin2023}.

This paper aims to bridge the disparity between the existing abilities of robots and the requirements posed by real-world scenarios, proposing a novel approach towards developing simple yet effective and intuitive robotic skill programming that does not necessitate specialized control expertise for application: visuo-tactile exploration of unknown rigid 3D curvatures through vision-augmented unified force-impedance control (VA-UFIC). By seamlessly integrating tactile and vision data to span various contact shapes between the tool and the environment, we develop a robust online contact alignment monitoring system, considering factors, e.g., tactile error, local surface curvature, and surface orientation. This information is seamlessly integrated into a vision-augmented unified force-impedance control framework, enabling the adjustment of robot stiffness and force regulation while exploring unknown rigid 3D curvatures. Visuo-tactile exploration is the next step to completing a force-motion planning framework that outputs an object-centric force-motion profile for an arbitrary tactile skill policy.

The contributions of this work include:
\begin{itemize}
\item[I] The introduction of online contact alignment monitoring to include various contact shapes between the tool and the environment: combining tactile error, the contact surface's local curvature, and surface orientation derived from tactile and vision data.
\item[II] Visuo-tactile exploration of unknown rigid 3D curvatures: integration of contact alignment monitoring into vision-augmented unified force-impedance control to adapt the robot's stiffness and regulate the force profile.
\item[III] Implementing virtual energy tanks to ensure system passivity and stability.
\item[IV] Evaluation of the proposed method's performance regarding contact alignment monitoring accuracy, real-time feedback latency, computational efficiency, and control performance using a Franka Emika research robot wiping challenging curvatures.
\end{itemize}
The remainder of the paper is organized as follows. Section~\ref{sec:method} presents the methodology, including visuo-tactile exploration of unknown rigid 3D curvatures through contact alignment monitoring using tactile data and vision. Additionally, it covers the passivity-based stability analysis for vision-augmented unified force-impedance control and the implementation of virtual energy tanks for stabilizing the system with variable stiffness and force regulation. The experimental protocol and corresponding results are detailed in Sections~\ref{sec:exp} and \ref{sec:results}, respectively. Finally, Section~\ref{sec:conc} provides the paper's conclusion.




\section{Methodology}\label{sec:method}
The methodology begins with designing and implementing unified force-impedance control, a well-established technique governing the robot's response to external forces while ensuring high compliance. This control framework integrates motion and force profiles to facilitate precise environmental interaction. Next, we explore the integration of tactile and vision inputs for contact alignment monitoring. This involves developing algorithms to interpret tactile data and vision cues to comprehensively understand the environment's geometry, e.g., curvatures. Using this sensory information as a foundation, we propose a visuo-tactile exploration of unknown rigid 3D curvatures by vision-augmented unified force-impedance control (VA-UFIC). This framework allows the robot to dynamically adjust its posture and modify stiffness, motion, and force policies to effectively respond to local faults during interactions with challenging surfaces. A thorough passivity-based stability analysis is conducted to ensure stability, identifying potential instabilities arising from variations in stiffness and force regulations. Additionally, we integrate virtual energy tanks into the control system to provide stability guarantees, particularly in the face of dynamic changes. 
\subsection{\bf{Control Design}}
For an n-DOF robot manipulator under unified force-impedance control during contact with gravity compensation, the Lagrangian dynamics is
\begin{align}
\label{eq:dyn}
   \bm{M}(\bm{q})\ddot{\bm{q}}+\bm{C}(\bm{q},\dot{\bm{q}})\dot{\bm{q}}+\bm{g}(\bm{q}) &= \bm{\tau}_\mathrm{c}+\bm{\tau}_\mathrm{ext}\,,
   \\
     \bm{\tau}_\mathrm{c} &= \bm{\tau}_\mathrm{i} + \bm{\tau}_\mathrm{f} + \bm{\tau}_\mathrm{g}\,,
\end{align}
where $\bm{\tau}_\mathrm{ext} \in \mathbb{R}^n$ represents the external torque exerted on the robot, while $\bm{M}(\bm{q}) \in \mathbb{R}^\mathrm{n \times n}$ denotes the robot mass matrix, $\bm{C}(\bm{q}, \dot{\bm{q}})\dot{\bm{q}} \in \mathbb{R}^\mathrm{n}$ signifies the Coriolis and centrifugal vector, and $\bm{g} \in \mathbb{R}^\mathrm{n}$ stands for the gravity vector in joint space. Additionally, $\bm{\tau}_\mathrm{c} \in \mathbb{R}^\mathrm{n}$ represents the control torque applied by the robot, which encompasses the torque command for controlling motion and force explicitly and separately, with $\bm{\tau}_\mathrm{g} \in \mathbb{R}^\mathrm{n}$ representing gravity compensation. Moreover, $\bm{\tau}_\mathrm{i}$ and $\bm{\tau}_\mathrm{f} \in \mathbb{R}^\mathrm{n}$ denote torques individually introduced by impedance and force control, respectively. Subsequently, we develop a control algorithm for the input torque $\bm{\tau}_\mathrm{c}$ to execute the desired tactile manipulation skill. This proposed control law for adaptive tactile skills extends from unified force-impedance control~\cite{Schindlbeck2015, Karacan2022}. Unified force-impedance control governs the robot's response to external forces, ensuring compliance while following motion and force profiles separately and explicitly. Starting with the robot's dynamics equation in Cartesian space
\begin{align}
\bm{M}_\mathrm{C}\ddot{\bm{x}} + \bm{C}_\mathrm{C}\dot{\bm{x}} +\bm{g}_\mathrm{C}&= \bm{f}_\mathrm{c}+\bm{f}_\mathrm{ext}\,,
\end{align}
where 
\begin{align}
\bm{M}_\mathrm{C} &= \bm{J}^\mathrm{\#T}\bm{M}\bm{J}^\mathrm{\#}\,,
\\
\bm{C}_\mathrm{C} &= \bm{J}^\mathrm{\#T}\bm{C}\bm{J}^\mathrm{\#}\,,
\\
\bm{g}_\mathrm{\mathrm{C}} &= \bm{J}^\mathrm{\#T}\bm{g}\,.
\end{align}
The external wrench to the base frame is denoted as $\bm{f}_\mathrm{ext} \in \mathbb{R}^6$. The robot mass matrix is represented as $\bm{M}_\mathrm{C}(\bm{q})$, where $\bm{q}$ is the joint configuration. The Coriolis and centrifugal effects are captured by $\bm{C}_\mathrm{C}(\bm{q}, \dot{\bm{q}}) \in \mathbb{R}^\mathrm{6 \times 6}$, and $\bm{g}_\mathrm{C}$ denotes the gravity vector in Cartesian space. Additionally, $\bm{f}_\mathrm{c}$ represents the wrench applied by the robot, which is related to the joint control torque $\bm{\tau}_\mathrm{c} \in \mathbb{R}^\mathrm{n}$ through the relationship $\bm{\tau}_\mathrm{c} = \bm{J}^\mathrm{T}(\bm{q}) \bm{f}_\mathrm{c}$, where $\bm{J} \in \mathbb{R}^\mathrm{6\times n}$ is the robot Jacobian matrix, and $\bm{J}^\#$ is the pseudo-inverse of the Jacobian. Compliance control, a subset of impedance control, omits inertia shaping and consequently excludes feedback of external forces. The compliance behavior is characterized by a time-varying stiffness matrix $\bm{K}_\mathrm{C}(t) \in \mathbb{R}^\mathrm{6\times6}$ and damping behavior determined by a positive definite matrix $\bm{D}_\mathrm{C} \in \mathbb{R}^\mathrm{6\times6}$. Moreover, $\bm{x} \in \mathbb{R}^6$ denotes the current pose of the end-effector in the base frame, and the pose error is denoted by $\tilde{\bm{x}}$. A conventional compliance controller for motion tracking can be formulated as
\begin{align}
\tilde{\bm{x}} &= \bm{x}-\bm{x}_\mathrm{d}\,,
\\
    \bm{f}_\mathrm{i} &= -\bm{K}_\mathrm{C}(t)\tilde{\bm{x}}-\bm{D}_\mathrm{C}\dot{\bm{x}}\,,
    \\
    \bm{\tau}_\mathrm{i} &= \bm{J}^T  \bm{f}_\mathrm{i}\,.
\end{align}

The force control is established to maintain the target contact force in the task space $\bm{f}^\mathrm{ee}_\mathrm{d} \in \mathbb{R}^6$, exerted by the robot concerning the external force $\bm{f}^\mathrm{ee}_\mathrm{\mathrm{ext}} \in \mathbb{R}^6$, as follows:
\begin{align}
    \bm{\tau}_\mathrm{f} &=  \bm{J}(\bm{q})^\mathrm{T} \bm{f}_\mathrm{f}, 
    \label{eq:frc} 
    \\
    \nonumber
    \bm{f}_\mathrm{f} &= \begin{bmatrix} [\bm{R}_\mathrm{ee}^\mathrm{0}]_\mathrm{3\times3} &  \bm{0}_\mathrm{3\times3} \\ \bm{0}_\mathrm{3 \times 3} & [\bm{R}_\mathrm{ee}^\mathrm{0}]_\mathrm{3\times3}
    \end{bmatrix} (\bm{f}_\mathrm{d}^\mathrm{ee} + \bm{K}_p \ \tilde{\bm{f}}_\mathrm{ext}^\mathrm{ee} + \\& \bm{K}_\mathrm{i} \int^t_0\tilde{\bm{f}}_\mathrm{ext}^\mathrm{ee} \ d\sigma)\, , \\
    \tilde{\bm{f}}^\mathrm{ee}_\mathrm{ext} &= \bm{f}^\mathrm{ee}_\mathrm{\mathrm{ext}}-\bm{f}^\mathrm{ee}_\mathrm{d}\, , 
\end{align}
In this context, $\bm{f}_\mathrm{f}$ $\in\mathbb{R}^6$ represents a feed-forward and feedback force controller in the base frame, which has been rotated by $\bm{R}_\mathrm{ee}^\mathrm{0}$. The proportional-integral (PI) controller gains are defined by the diagonal matrices $\bm{K}_\mathrm{p}$ and $\bm{K}_\mathrm{i}\in$ $\mathbb{R}^\mathrm{6\times6}$. The resultant control torque without the gravity compensation for unified force-impedance control $\bm{\tau} \in \mathbb{R}^n$ is
\begin{align}
        \bm{\tau} &=\bm{\tau}_\mathrm{f} + \bm{\tau}_\mathrm{i}\,.
\end{align}
Next, we introduce contact alignment monitoring based on visual and tactile data to explore unknown rigid 3D curvatures for an arbitrary force-motion policy in the end-effector frame. This rich sensory information is augmented to unified force-impedance control so that the control parameters, such as stiffness and contact force shaping function, are decided. Thus, the robot can maintain contact with the current surface geometry and orientation.

\subsection{\bf{Visuo-Tactile Exploration of Unknown Rigid 3D Curvatures by VA-UFIC}}
To guarantee a successful execution of the desired skill and to understand the environment comprehensively, we monitor the contact alignment that utilizes tactile and visual perception. Based on this rich sensory information, we enable the robot to adjust posture, stiffness, motion, and force policy for local fault recovery during interactions with challenging surfaces at the low-level control.

The visual perception algorithm operates through two concurrent threads: (i) data collection and pre-processing and (ii) surface normal estimation. Initially, depth images are transformed into point clouds for use within the Point Cloud Library~\cite{rusu}. RGB and depth images are acquired from the video stream, precisely aligned, and converted into a 3D point cloud. Subsequently, a surface normal estimation method is applied based on the acquired point cloud data to predict contact surface orientation. Inspired by Westfechtel et al.~\cite{thomas}, the region growing method clusters the surface normals within similar orientations to segment the point clouds. Principal Component Analysis (PCA) is then employed on the clustered segments to determine their orientation.

The eigenvectors of the covariance matrix $\bm{\Sigma}_\mathrm{3\times3} = [\bm{e}_\mathrm{1}; \bm{e}_\mathrm{2}; \bm{n}^\mathrm{c}_\mathrm{s}]$, representing the PCA output, characterize a segment's primary directions. Here, $\bm{n}^\mathrm{c}_\mathrm{s} \in \mathbb{R}^{3\times3}$ denotes the surface normal of the segment in the camera frame, which represents the direction of a surface segment, while $\bm{e_}\mathrm{1}$ and $\bm{e}_\mathrm{2}$ represent the long and short edges, respectively. Furthermore, the local curvature $l_\mathrm{s}$ of the working surface can be computed using Equation \eqref{Eq:localcurvature}, where $\lambda_\mathrm{i}\,, i = 1,2,3$ are the eigenvalues of the covariance matrix $\bm{\Sigma}$ obtained through PCA. 
\begin{align}
	\label{eq:surface normal}
    \bm{n} &= [0 \, 0 \, 1]^\mathrm{T} \, ,
    \\
    \mathrm{\theta} &= |\mathrm{cos}^\mathrm{-1}(\bm{n}^{\mathrm{c}\mathrm{T}}_\mathrm{s}\bm{n})|\, ,
     \label{eq:visual_error}
      \\
     \label{Eq:localcurvature}
     l_\mathrm{s} &=\bigg |\frac{\lambda_3 }{tr(\bm{\Sigma})}\bigg |\,.
\end{align}
The end-effector orientation error $\theta$, representing the deviation in surface normal between $\bm{n}^\mathrm{c}_\mathrm{s}$ captured by the camera and the z-axis of the camera (aligned with z-axis of the end-effector), can be determined using the $\mathrm{acos}$ function in~\eqref{eq:visual_error}. Undesired contacts lead to deviations from the desired pose, manifesting as either a pose error $\tilde{\bm{x}}^\mathrm{ee}\in \mathbb{R}^6$ or external forces $\bm{f}^\mathrm{ee}_\mathrm{ext} \in \mathbb{R}^\mathrm{6}$ at the end-effector. In real-time, contact alignment monitoring accumulates all the error terms and their corresponding signal strengths as presented in~\eqref{eq:error}. Simultaneously, the signal strengths for the tactile error, surface normal deviation, and local curvature term, denoted as $\mathrm{\alpha}$, $\mathrm{\xi}$, and $\mathrm{\gamma}$ respectively, contribute to the adaptive process and decide how agile the robot reacts to them. In~\eqref{eq:metric}, the contact alignment monitoring $\mathrm{C}$ is employed to calculate a normalized coefficient $\mathrm{h}$. 
\begin{align}
	\label{eq:error}
    \mathrm{C} &= | \alpha|\bm{f}^{\mathrm{ee}
\mathrm{T}}_\mathrm{ext} \tilde{\bm{x}}^\mathrm{ee}| + \xi\theta + \gamma{l}_\mathrm{s}|\,,
    \\
	\mathrm{h} &= 1-\frac{\mathrm{C}}{\mathrm{C}_\mathrm{\mathrm{m}}} \, .   
 \label{eq:metric}
\end{align}

The contact alignment margin $\mathrm{C}_\mathrm{m}$ is crucial for compensating for minor environmental effects, such as surface friction and measurement inaccuracy, and, notably, employing position rather than velocity or acceleration results in a less noisy signal. The normalized metric $h$ is subsequently linked to the maximum stiffness level at the end-effector frame $\bm{K}^\mathrm{ee}_\mathrm{max,t}$ through $\rho_\mathrm{align}$ and it is rotated back to the base frame by the rotation matrix $\bm{R}_\mathrm{ee}^\mathrm{0}$. This inherent behavior is leveraged to robustly respond to undesired contacts and reconfigure the end-effector through adaptive adjustments to the stiffness matrix in the translational directions $\bm{K}_\mathrm{C,t}$.
\begin{align}
	\bm{K}_\mathrm{C,t}&= \rho_\mathrm{align}\bm{R}_\mathrm{ee}^\mathrm{0}\bm{K}^\mathrm{ee}_\mathrm{max,t}\bm{R}_\mathrm{ee}^{\mathrm{0} \mathrm{T}} \,. 
	\label{eq:stiffness}
\end{align}
The alignment parameter $\rho_\mathrm{align}$ is extended based on studies~\cite{Karacan2022, Karacan2023}, as outlined in~\eqref{eq:rhoDot}.
\begin{align}
    \dot{\rho}_\mathrm{align} &= 	
    \begin{cases}
    \min\{\rho,0\} \, , & \rho_\mathrm{align} = 1
    \\
   \rho \, , & 0 < \rho_\mathrm{align} < 1,\,\, \rho_\mathrm{align}(0)=0,
   \\
 \max \{\rho,0\} \, , & \rho_\mathrm{align} = 0
\end{cases}
\label{eq:rhoDot}
\end{align}
and $\rho$ is given by
\begin{align}
	\rho&=\mathrm{h}\rho_\mathrm{align} + {\rho_\mathrm{min}}\, .
 \label{eq:rho}
\end{align}
It's important to note that, to ensure an initial increment when $\rho_\mathrm{align} = 0$, a small positive constant ${\rho_\mathrm{min}}$ is introduced into the dynamics of the shaping function. When the robot is entirely compliant, meaning $\rho_\mathrm{align}$ equals zero, it becomes capable of adapting to the environment. This implies that the translational component of the actual end-effector pose $\bm{x}_\mathrm{ee,t} \in \mathbb{R}^3$ is fed back to the controller as the desired translational pose $\bm{x}_\mathrm{d,t}$. Subsequently, we calculate the rotation of the end-effector $\bm{R}^\mathrm{0}_\mathrm{ee,d} \in \mathbb{R}^{3\times3}$ to adapt to the environment.

Upon detecting a significant deviation from the intended contact alignment between the robotic tool and the surface, often resulting from abrupt changes in the contact, the robot becomes compliant in translational directions. Afterward, it realigns itself with the detected surface normal in the camera frame, denoted as $\bm{n}^c_\mathrm{s}$, and regenerates the motion and force policy. This process necessitates the knowledge of the desired end-effector orientation $\bm{R}^\mathrm{0}_\mathrm{ee,d}$, which can be computed from the detected surface normal $\bm{n}^c_\mathrm{s}$. The contact surface normal is initially transformed from the camera frame to the end-effector frame, then to the robot base frame using~\eqref{eq:SN2base}. $\bm{R}^\mathrm{ee}_\mathrm{c}$ represents the rotation matrix that transfers from the camera frame to the end-effector frame. Similarly, $\bm{R}^\mathrm{0}_\mathrm{c}$ denotes the rotation matrix that transfers from the camera frame to the robot base frame. Subsequently, the surface normal $\bm{n}^\mathrm{0}_\mathrm{s}$ in the base frame contributes to the construction of the desired orientation matrix $\bm{R}^\mathrm{0}_\mathrm{ee,d}$ through the following steps: i.) read the current orientation of the end-effector and extract the first column $\bm{r}_\mathrm{x}$; ii.) project $\bm{r}_\mathrm{x}$ onto the orthogonal plane of the surface normal, as per~\eqref{eq:XProjection}; iii.) calculate the second column $\bm{r}_\mathrm{y}$ through the cross product of the projected $\bm{r}^\mathrm{'}_\mathrm{x}$ and surface normal $\bm{n}^\mathrm{0}_\mathrm{s}$; iv.) assemble these three distinct components into the desired rotational matrix $\bm{R}^\mathrm{0}_\mathrm{ee,d}$.
\begin{align}
    \label{eq:SN2base}
    \bm{R}^\mathrm{0}_\mathrm{c} &= \bm{R}^\mathrm{0}_\mathrm{ee}\bm{R}^\mathrm{ee}_\mathrm{c}\, ,
    \bm{n}^\mathrm{0}_\mathrm{s} = \bm{R}^\mathrm{0}_\mathrm{c}\bm{n}^\mathrm{c}_\mathrm{s}\, ,
    \\
    \label{eq:readCurrentR}
    \bm{R}^\mathrm{0}_\mathrm{ee} &= \begin{bmatrix} [\bm{r}_\mathrm{x}]_\mathrm{3\times1} &   [\bm{r}_\mathrm{y}]_\mathrm{3\times1} & [\bm{r}_\mathrm{z}]_\mathrm{3\times1}
    \end{bmatrix}\, , 
    \\
    \label{eq:ZProjection}
    \bm{r}_\mathrm{z}^\mathrm{\prime} &= \bm{n}^\mathrm{0}_\mathrm{s} ,
    \\
    \label{eq:XProjection}
    \bm{r}^\mathrm{'}_\mathrm{x} &= \bm{r}_\mathrm{x} - (\bm{r}_\mathrm{x}^\mathrm{T}\bm{n}^\mathrm{0}_\mathrm{s})\bm{n}^\mathrm{0}_\mathrm{s}\, ,
    \\
    \label{eq:YProjection}
    \bm{r}^\mathrm{'}_\mathrm{y} &= \bm{r}^\mathrm{'}_\mathrm{z} \times \bm{r}^\mathrm{'}_\mathrm{x}\, ,
    \\
    \label{eq:SN2matrix}
     \bm{R}^\mathrm{0}_\mathrm{ee,d} &= \begin{bmatrix} [\bm{r}^\mathrm{'}_\mathrm{x}]_\mathrm{3\times1} &   [\bm{r}^\mathrm{'}_\mathrm{y}]_\mathrm{3\times1} & [\bm{r}^\mathrm{'}_\mathrm{z}]_\mathrm{3\times1}
        \end{bmatrix}\, .
\end{align}
To ensure that the input signal provided to the robot is smooth and continuous, a low-pass filter is implemented, facilitating the gradual transition of the signal $\bm{R}^\mathrm{0}_\mathrm{\mathrm{input}}$ from the initial rotation $\bm{R}^\mathrm{0}_\mathrm{init}$ to the desired rotation $\bm{R}^\mathrm{0}_\mathrm{ee,d}$. The scaling coefficient $\zeta$ falls within the range of 0 to 1, and $T$ represents the time interval governing the convergence of the low-pass filter. Specifically, at $t = 0$, signifying the initiation of contact alignment, the output is the original rotation $\bm{R}^\mathrm{0}_\mathrm{init}$. Conversely, when $t = T$, indicating the completion of convergence, the input rotation to the robot becomes $\bm{R}^\mathrm{0}_\mathrm{ee,d}$.
\begin{align}
    \label{eq:zeta}
    \zeta &= \frac{t}{T}\,, 0\le t \le T\, ,
    \\
    \label{eq:lowpassfilter}
    \bm{R}^\mathrm{0}_\mathrm{input} &= (\bm{R}^\mathrm{0}_\mathrm{ee,d}\bm{R}^{\mathrm{0}\mathrm{T}}_\mathrm{init})^\mathrm{\zeta}\bm{R}^\mathrm{0}_\mathrm{init}\, .
\end{align}
Furthermore, we formulate the force shaping function $\rho_\mathrm{frc}$. This function facilitates the alignment of the commanded force to compensate for tool alignment errors and mitigate the undesired loss of contacts. The robot accommodates the tool alignment error $\bm{f}^{\mathrm{ee}\mathrm{T}}_\mathrm{d}\tilde{\bm{x}}^\mathrm{ee}$ during contact loss within the error margin $\delta_\mathrm{c}>0$. Additionally, in cases where the robot loses surface contact due to a substantial tool alignment error, it transitions to impedance control, adhering solely to the desired motion.
\begin{align}
    \rho_\mathrm{frc} \hspace{-0.5mm} &= \begin{cases}
    1 \, , & \bm{f}^{\mathrm{ee}\mathrm{T}}_\mathrm{d}\tilde{\bm{x}}^\mathrm{ee}\leq 0
    \\
    0.5 (1 + \cos( (\pi\frac{\tilde{x}^\mathrm{ee}_\mathrm{z}}{\delta_\mathrm{c}}))) \, , & 0<\bm{f}^{\mathrm{ee}\mathrm{T}}_\mathrm{d}\tilde{\bm{x}}^\mathrm{ee} \land \\ & 0< \tilde{x}^\mathrm{ee}_\mathrm{z}\leq \delta_\mathrm{c} 
    \\
    0 \, , &  \mathrm{otherwise}.
    \end{cases}
\end{align}
\begin{align}
\bm{\tau}_\mathrm{f}&=\rho_\mathrm{frc}\bm{J}^\mathrm{T}\bm{f}_\mathrm{f}\, .
\end{align}

Finally, after calculating the orientation matrix, the desired trajectory and force policy can be adapted accordingly in the base frame. However, stiffness variation in the impedance controller and, in case of loss of contact, the force controller may jeopardize the controller's stability, resulting in unsafe behaviour~\cite{Kronander2016}. To address it, the virtual energy tanks are used for variable stiffness and force regulation to ensure the system's passivity and stability.

\subsection{\bf{Passivity-Based Stability Analysis and Installing Virtual Energy Tanks}}
Virtual energy tanks are integrated to guarantee stability by identifying potential instabilities arising from stiffness variations and force regulations to ensure stability even amidst dynamic changes~\cite{Stramigioli2015,Michel2020}. To show the passivity, the storage function $S_\mathrm{r}$ for the Cartesian robot dynamics that represents the kinetic energy of the robot is 
\begin{align}
    S_\mathrm{r} &= \frac{1}{2}\dot{\bm{x}}^T\bm{M}_\mathrm{C}\dot{\bm{x}}\,,
\end{align}
where the time derivative of the storage function $\dot{S}_\mathrm{r}$ is
\begin{align}
        \dot{S}_\mathrm{r} &= \dot{\bm{x}}^T(\bm{f}+\bm{f}_\mathrm{ext})\, \ \text{or}  \   \dot{S}_\mathrm{r}= \dot{\bm{q}}^T(\bm{\tau}+\bm{\tau}_\mathrm{ext})\, , 
\end{align}
which we can say that it is passive for the pair ($\bm{\tau}+\bm{\tau}_\mathrm{ext},\dot{\bm{q}}$). Identifying potential instabilities arising from stiffness variations and force regulations, we split the problem of analyzing the stability of robot dynamics into two cases: without contact (Case I) and during contact (Case II). 

\subsubsection{\textbf{Case I: Stability analysis without any contact}} 
When there is no contact, the stiffness $\bm{K}_\mathrm{C}$ remains constant. Thus, only the force controller may cause instability. The stability of the force controller can be assessed using the subsequent storage function $S_\mathrm{f}$
\begin{align}
S_\mathrm{f} &=\frac{1}{2}\tilde{\bm{x}}^T\bm{K}_\mathrm{C}\tilde{\bm{x}}\,,
\\
\dot{S}_\mathrm{f} &=\dot{\bm{x}}^T\bm{K}_\mathrm{C}\tilde{\bm{x}}\,,
\\
&=\dot{\bm{x}}^T(-\bm{f}+\bm{f}_\mathrm{f}-\bm{D}_\mathrm{C}\dot{\bm{x}})\,.
\end{align}
Due to the pair of ($\bm{f}_\mathrm{f},\dot{\bm{x}}$), the force controller should be modified to guarantee stability by augmenting a virtual energy tank such that 
\begin{align}
\bm{f}&=-\bm{D}_\mathrm{C}\dot{\bm{x}}-\bm{K}_\mathrm{C}\tilde{\bm{x}}+\lambda\bm{f}_\mathrm{f}+\bm{f}_\mathrm{f,var}\,.
\end{align}
To indicate if the force controller $\bm{f}_\mathrm{f}$ is passive, $\lambda$ is used such that
\begin{align}
       \lambda&=  \begin{cases} 1, & \dot{\bm{\bm{x}}}^T\bm{f}_\mathrm{f}<0\,,
    \\
    0, & else
    \end{cases}
\end{align}
Moreover, $\bm{f}_\mathrm{f,var}$ is the modification in the controller regulated by the tank. Being the tank's energy is $ S_\mathrm{t,f}$, we design its dynamics $\dot{x}_\mathrm{t,f}$ as
\begin{align}
    \dot{x}_\mathrm{t,f} &=\lambda\beta_\mathrm{f}\frac{\dot{\bm{\bm{x}}}^T\bm{D}_\mathrm{C}\dot{\bm{x}}-\dot{\bm{\bm{x}}}^T\bm{f}_\mathrm{f}}{x_\mathrm{t,f}}+u_\mathrm{t,f}\,,
    \\ 
    y_\mathrm{t,f} &= x_\mathrm{t,f}\,,
    \\
    S_\mathrm{t,f} &=\frac{1}{2}x^2_\mathrm{t,f}\,,
\end{align}
where $x_\mathrm{t,f}$,  $u_\mathrm{t,f}$, and  $y_\mathrm{t,f}$  are the tank's state, input, and out variable, respectively. The tank is interconnected to the controllers through the power-preserving Dirac structure
\begin{align}
    \begin{bmatrix}
        \bm{f}_\mathrm{f,var} \\ u_\mathrm{t,f} 
        \end{bmatrix} &= \begin{bmatrix} 0 & \bm{\omega}_\mathrm{f} \\ -\bm{\omega}_\mathrm{f}^T & 0
    \end{bmatrix}\begin{bmatrix}
        \dot{\bm{x}} \\ y_\mathrm{t,f}
    \end{bmatrix}\,,
\\
    \bm{\omega}_\mathrm{f} &=\frac{\sigma(S_\mathrm{t,f})(1-\lambda)\bm{f}_\mathrm{f}}{y_\mathrm{t,f}}\,.
\end{align}
The design parameter $\bm{\omega}_\mathrm{f}$ is a modulating factor that controls the power transmission between the tank and the controller with the valve $\sigma(S_\mathrm{t,f})$ is
\begin{align}
       \sigma(S_\mathrm{t,f})&=  \begin{cases} \sigma(S_\mathrm{t,f}) \in (0,1], & S_\mathrm{t,f}>\underline{S}_\mathrm{t,f}\,,
    \\
    0, & else
    \end{cases}
\end{align}
The controller can regulate force if the tank is not depleted. Note that, to avoid singularities, we set a lower limit $\underline{S}_\mathrm{t,f}$ for the energy threshold in the tank. Additionally, to ensure the tank is not overloaded, a specific upper-limit $\overline{S}_\mathrm{t,f}$ for the tank is introduced:
\begin{align}
    \beta_\mathrm{f}&=  \begin{cases} \kappa_\mathrm{f} \in [0,1], & S_\mathrm{t,f}<\overline{S}_\mathrm{t,f}\,,
    \\
    0, & else
    \end{cases}
\end{align}
where a smooth transition behavior $ \kappa_\mathrm{f}$ is embedded. Using $\bm{f}_\mathrm{f,var}=\bm{\omega}_\mathrm{f} y_\mathrm{t,f}$, the passivity of the subsystem  $S_\mathrm{c}$ involving the tank and the controller with the combined storage function $S_\mathrm{overall}$
\begin{align}
        S_\mathrm{overall}&= S_\mathrm{c}+S_\mathrm{r}\,, \ S_\mathrm{c} = S_\mathrm{f} + S_\mathrm{t,f}\,,
        \\
    \dot{S}_\mathrm{overall} &= \dot{S}_\mathrm{c} +\dot{S}_\mathrm{r}\,,
        \\ \nonumber
     &= -\dot{\bm{x}}^T\bm{f}+\lambda(1-\beta_\mathrm{f})\dot{\bm{x}}^T\bm{f}_\mathrm{f}- \\&  (1-\lambda\beta_\mathrm{f})\dot{\bm{x}}^T\bm{D}_\mathrm{C}\dot{\bm{x}}+\dot{\bm{x}}^T(\bm{f}+\bm{f}_\mathrm{ext})\,,
   \\
     &= \dot{\bm{q}}^T\bm{\tau}_\mathrm{ext}-(1-\lambda\beta_\mathrm{f})\dot{\bm{\bm{q}}}^T\bm{J}^T\bm{D}_\mathrm{C}\dot{\bm{x}}+\\ &\lambda(1-\beta_\mathrm{f})\dot{\bm{q}}^T\bm{\tau}_\mathrm{f}\,.
\end{align}
The modified unified force-impedance control ensures stability in case of loss of contact:
\begin{align}
\bm{f}&=-\bm{D}_\mathrm{C}\dot{\bm{x}}-\bm{K}_\mathrm{C}\tilde{\bm{x}}+\rho_\mathrm{frc}(\lambda+\sigma(S_\mathrm{t,f})(1-\lambda))\bm{f}_\mathrm{f}\,.
\end{align}

\subsubsection{\textbf{Case II: Stability analysis during contact} }
Contact means that while the robot can move in k-dimensions, the motion is constrained in the rest $6-k$ dimensions. Thus, the force controller during contact does not jeopardize stability, as the components of $\dot{\bm{x}}$ in the force control direction is zero $\dot{\bm{x}}^T\bm{f}_\mathrm{f}=0$. However, stiffness variation in the impedance controller during contact may cause instability. Next, we present the stability analysis and the virtual energy tank for the impedance controller. To assess the passivity of this controller, we examine the storage function, which is regarded as the corresponding spring potential $S_\mathrm{i}$
\begin{align}
    S_\mathrm{i} &= \frac{1}{2}\tilde{\bm{x}}^T\bm{K}_\mathrm{C}\tilde{\bm{x}}\,,
    \\
    \dot{S}_\mathrm{i} &= \dot{\bm{x}}^T\bm{K}_\mathrm{C}\tilde{\bm{x}}+\frac{1}{2}\tilde{\bm{x}}^T\dot{\bm{K}_\mathrm{C}}\tilde{\bm{x}}\,,
    \\
    &=-\dot{\bm{x}}^T\bm{f}_\mathrm{i}-\dot{\bm{\bm{x}}}^T\bm{D}_\mathrm{C}\dot{\bm{x}}+\frac{1}{2}\tilde{\bm{x}}^T\dot{\bm{K}_\mathrm{C}}\tilde{\bm{x}}\,.
\end{align}
Passivity with respect to the pair $(\bm{f}_\mathrm{i},\dot{\bm{x}})$ cannot be guaranteed due to the term of $\frac{1}{2}\tilde{\bm{x}}^T\dot{\bm{K}}\tilde{\bm{x}}$. Therefore, we modify the impedance controller $\bm{f}_\mathrm{i}$ by adding a control term $\bm{f}_\mathrm{i,var}$ regulated by the energy tank
\begin{align}
    \bm{f}_\mathrm{i} &=-\bm{K}_\mathrm{const}\tilde{\bm{x}}-\bm{D}_\mathrm{C}\dot{\bm{x}}+\bm{f}_\mathrm{i,var}\,,
    \\
\bm{K}_\mathrm{C} &= \bm{K}_\mathrm{const} + \bm{K}_\mathrm{var}(t)\,,
\end{align}
where the stiffness matrix has constant $ \bm{K}_\mathrm{const}$, which can also be zero, and time-varying parts $ \bm{K}_\mathrm{var}(t)=\rho_\mathrm{align}\bm{K}_\mathrm{max}$. Energy tank state $x_\mathrm{t,i}$,  its dynamics $\dot{x}_\mathrm{t,i}$, and the tank energy $S_\mathrm{t,i}$ are 
\begin{align}
    \dot{x}_\mathrm{t,i} &= \beta_\mathrm{i}\frac{\dot{\bm{\bm{x}}}^T\bm{D}\dot{\bm{x}}}{x_\mathrm{t,i}}+u_\mathrm{t,i}\,,
    \\
y_\mathrm{t,i} &= x_\mathrm{t,i}\,,
\\
S_\mathrm{t,i} &= \frac{1}{2}x^2_\mathrm{t,i}\,,
\end{align}
where $u_\mathrm{t,i}$ and $y_\mathrm{t,i}$ are input and output variable, respectively. To ensure the tank is not overloaded, a specific upper-limit $\overline{S}_\mathrm{t,i}$ for the tank is introduced with a smooth transition behavior $ \kappa_\mathrm{i}$
\begin{align}
    \beta_\mathrm{i}&=  \begin{cases} \kappa_\mathrm{i} \in [0,1], & S_\mathrm{t,i}<\overline{S}_\mathrm{t,i}\,,
    \\
    0, & else
    \end{cases}
\end{align}
The Dirac structure for the ports implies the passivity of the system:
\begin{align}
    \begin{bmatrix}
        \bm{f}_\mathrm{i,var} \\ u_\mathrm{t,i} 
        \end{bmatrix} &= \begin{bmatrix} 0 & \bm{\omega}_\mathrm{i} \\ -\bm{\omega}_\mathrm{i}^T & 0
    \end{bmatrix}\begin{bmatrix}
        \dot{\bm{x}} \\ y_\mathrm{t,i}
    \end{bmatrix}\,,
\\
    \bm{\omega}_\mathrm{i} &=-\frac{\sigma(S_\mathrm{t,i})\bm{K}_\mathrm{var}(t)\tilde{\bm{x}}}{y_\mathrm{t,i}}\,.
\end{align}
The design parameter $\bm{\omega}_\mathrm{i}$ is a modulating factor that controls the power transmission between the tank and the impedance controller with the valve $\sigma(S_\mathrm{t,i})$
\begin{align}
       \sigma(S_\mathrm{t,i})&=  \begin{cases} \sigma(S_\mathrm{t,i}) \in (0,1], & S_\mathrm{t,i}>\underline{S}_\mathrm{t,i}\,,
    \\
    0, & else
    \end{cases}
\end{align}
This indicates that the controller can adjust stiffness if the tank has not been depleted. To prevent singularities, we establish a lower limit, denoted as $\underline{S}_\mathrm{t,i}$, for the energy threshold in the tank. Furthermore, ensuring the passivity of subsystem $S_\mathrm{c}$, which comprises the tank and the controller, is achieved by combining the storage function in the following:
\begin{align}
    S_\mathrm{c} &= \frac{1}{2}\tilde{\bm{x}}^T\bm{K}_\mathrm{const}\tilde{\bm{x}} +  \frac{1}{2}x^2_\mathrm{t,i}\,, \\
\dot{S}_\mathrm{c} &= \dot{\bm{x}}\bm{K}_\mathrm{const}\tilde{\bm{x}}+\dot{x}_\mathrm{t,i}x_\mathrm{t,i}\,.
\end{align}
Using $\bm{f}_\mathrm{i,var}=\bm{\omega}_\mathrm{i} y_\mathrm{t,i}$
\begin{align}
     \dot{S}_\mathrm{c} &= -\dot{\bm{x}}^T\bm{f}_\mathrm{i}-\dot{\bm{\bm{x}}}^T\bm{D}_\mathrm{C}\dot{\bm{x}}+\dot{\bm{x}}^T\bm{\omega}_\mathrm{i} y_\mathrm{t,i}+ \\& \beta_\mathrm{i}\dot{\bm{\bm{x}}}^T\bm{D}_\mathrm{C}\dot{\bm{x}}-\bm{\omega}_\mathrm{i}^T\dot{\bm{x}}x_\mathrm{t,i}\,,
     \\
 &= -\dot{\bm{q}}^T \bm{\tau}_\mathrm{i}-(1-\beta_\mathrm{i})\dot{\bm{\bm{x}}}^T\bm{D}_\mathrm{C}\dot{\bm{x}}\,.
\end{align}
The total storage function $ S_\mathrm{overall}$ and its time derivative $  \dot{S}_\mathrm{overall}$ are 
\begin{align}
    S_\mathrm{overall} &= S_\mathrm{c} +S_\mathrm{r}\, ,
    \dot{S}_\mathrm{overall} = \dot{S}_\mathrm{c} +\dot{S}_\mathrm{r}\, ,
    \\
     \dot{S}_\mathrm{overall} &= -\dot{\bm{q}}^T \bm{\tau}-(1-\beta_\mathrm{i})\dot{\bm{\bm{x}}}^T\bm{D}_\mathrm{C}\dot{\bm{x}}+ \dot{\bm{q}}^T(\bm{\tau}+\bm{\tau}_\mathrm{ext})\,, \\ &= \dot{\bm{q}}^T\bm{\tau}_\mathrm{ext}-(1-\beta_\mathrm{i})\dot{\bm{\bm{q}}}^T\bm{J}^\mathrm{T}\bm{D}_\mathrm{C}\dot{\bm{x}}\,.
\end{align}
Finally, the modified unified-impedance control ensures stability during contact and no-contact:
\begin{align}
    \bm{f} &= -\bm{K}_\mathrm{const}\tilde{\bm{x}}-\bm{D}_\mathrm{C}\dot{\bm{x}}-\sigma(S_\mathrm{t,i})\rho_\mathrm{align}\bm{K}_\mathrm{max}\tilde{\bm{x}}+\\& \rho_\mathrm{frc}(\lambda+\sigma(S_\mathrm{t,f})(1-\lambda))\bm{f}_\mathrm{f}\,.
\end{align}
Next, the validation scenarios and relevant performance metrics for the exemplary tactile skill to explore the curvatures are discussed. 

\section{Experimental Validation} \label{sec:exp}
To assess our framework's visuo-tactile exploration and control performance for exploring an unknown rigid curvature, we conduct experiments using a Franka Emika robot to perform a wiping policy. We employ an Intel RealSense D435i camera (Intel Corp., USA) to capture environmental information. The camera is positioned at the robot's flange to minimize body occlusion, and its axis aligns with the z-axis of the end-effector frame during execution. This alignment simplifies the transformation from the task frame to the base frame, enhancing the accuracy of surface normal estimation. The visual pipeline and the robot's master controller run on a mini-ITX PC (HP Z2 Mini G5 Workstation with Intel i7-10700t). The contact surface is 3D-printed with dimensions of $0.26\times0.51\times h$, where $h=0.02\sin(\frac{\pi}{0.19}y+0.44)+0.02$ \SI{}{m}. 

The experimental procedure evaluates the accuracy of contact alignment monitoring, real-time feedback latency, computational efficiency, and control performance while exploring the unknown 3D rigid curvature for an arbitrarily given wiping policy.
\begin{table}[t]
\caption{Parameters used in the experiments.}           \vspace{-0.2cm} 
\label{tab:parameters}
\centering
\begin{tabular}{|c|c|c|}  
\hline
Parameter     & Unit    & Value   \\   \hline
$\bm{K}_\mathrm{max}$    & N/m & diag[1000,1000,10,200,200,200] \\ 
damping coefficient      & -    & diag[0.7,0.7,0.7,1,1,1]      \\ 
$C_\mathrm{m}$           & -    & 0.9       \\ 
$\alpha$ \,, $\xi$ \,, $\gamma$ & - & 1\,, 0.08\,, 10 \\ 
$\bm{K}_p$, $\bm{K}_i$ & - & 0.6$\bm{I}_\mathrm{6\times6}$, 0.3$\bm{I}_\mathrm{6\times6}$ \\
$\delta_\mathrm{c}$ &m &0.04 \\ 
$\rho_\mathrm{min}$      &-    & 0.001 \\
$x_\mathrm{t,i}(0)$            &-    & 7       \\
$\overline{S}_\mathrm{t,i}$   & J   &32       \\
$\underline{S}_\mathrm{t,i}$  &J    & 1       \\
$x_\mathrm{t,f}(0)$   & -    & 2       \\
$\overline{S}_\mathrm{t,f}$ & J    & 2       \\ 
$\underline{S}_\mathrm{t,f}$  & J    & 1       \\
    \hline
\end{tabular}
\end{table}

\subsection{\bf{Experimental Procedure}}
During wiping, we conduct experiments for visuo-tactile exploration of the unknown, challenging curved surface. The arbitrary wiping tactile policy used during the exploration is 
\begin{align}
    \bm{f}^\mathrm{ee}_\mathrm{d} &= [0,0,15,0,0,0]\,,
\nonumber    \\
    \bm{x}^\mathrm{ee}_\mathrm{d}&=[0.04 \sin(2t), 0.04 (\cos(2t)-1)-0.005t,0,0,0,0 ]\,, \nonumber
\end{align}
where $t$ is the time parameter. Moreover, Table~\ref{tab:parameters} shows other parameters designed for the experiments. 

First, the robot starts without contacting and aligning to the surface. The expected behavior is that the robot aligns itself, establishes contact, and explores the curvatures while presenting passive and accurate force-motion tracking.

\subsection{\bf{Performance Metrics}}
Performance metrics employed for validation include i.) contact alignment monitoring accuracy, ii.) real-time feedback latency, iii.) computational efficiency, and iv.) control performance. Contact alignment monitoring accuracy measures the precision of exploring the surface being wiped. Real-time feedback latency quantifies the time delay between sensing contact with the surface and adjusting the robot's motion or force. Computational efficiency measures the processing rate achieved by the visual pipeline, considering factors such as frame rate and latency. Control performance encompasses quantifying the accuracy and precision of the robot's movements during the wiping task, measuring deviations from the desired wiping trajectory or force profile to assess control robustness, and evaluating control performance metrics such as root mean square and absolute mean error in tracking. Force control precision considers the uniformity of pressure applied during wiping, ensuring consistent cleaning or polishing without damaging delicate surfaces. It entails measuring the deviation between the desired and actual force exerted on the surface and assessing force control performance using root mean square error and absolute mean error.

\section{Results and Discussion} \label{sec:results}
\begin{figure}
\centering	
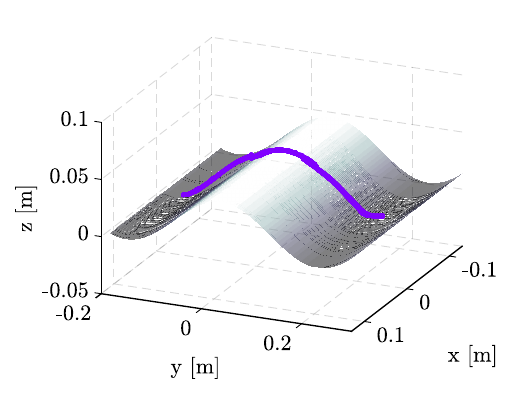
    \vspace{-0.8cm}
\caption{\textbf{Visuo-Tactile Exploration of an Unknown Rigid 3D Curvature.} The robot's actual trajectory along the y- and z-direction in the base frame is compared to the model of the contact surface.}
\label{results:3d}
\end{figure}

\begin{figure}
\centering	
\begingroup%
  \makeatletter%
  \providecommand\color[2][]{%
    \errmessage{(Inkscape) Color is used for the text in Inkscape, but the package 'color.sty' is not loaded}%
    \renewcommand\color[2][]{}%
  }%
  \providecommand\transparent[1]{%
    \errmessage{(Inkscape) Transparency is used (non-zero) for the text in Inkscape, but the package 'transparent.sty' is not loaded}%
    \renewcommand\transparent[1]{}%
  }%
  \providecommand\rotatebox[2]{#2}%
  \newcommand*\fsize{\dimexpr\f@size pt\relax}%
  \newcommand*\lineheight[1]{\fontsize{\fsize}{#1\fsize}\selectfont}%
  \ifx\svgwidth\undefined%
    \setlength{\unitlength}{255.11811024bp}%
    \ifx\svgscale\undefined%
      \relax%
    \else%
      \setlength{\unitlength}{\unitlength * \real{\svgscale}}%
    \fi%
  \else%
    \setlength{\unitlength}{\svgwidth}%
  \fi%
  \global\let\svgwidth\undefined%
  \global\let\svgscale\undefined%
  \makeatother%
  \begin{picture}(1,1.2)%
    \lineheight{1}%
    \setlength\tabcolsep{0pt}%
    \put(0,0){\includegraphics[width=\unitlength,page=1]{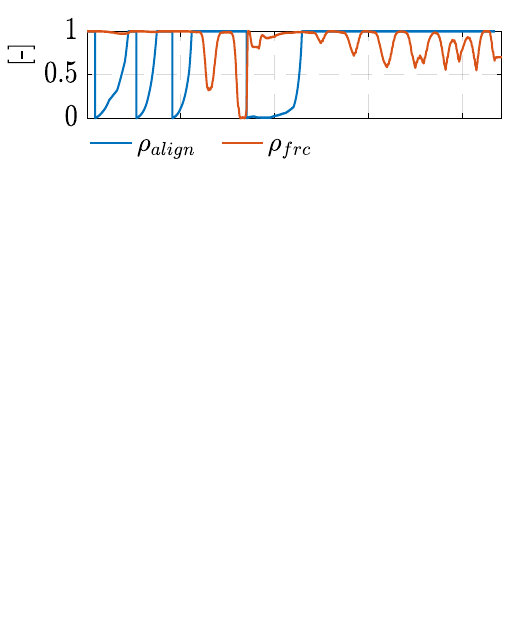}}%
    \put(0.01811425,1.15651659){\makebox(0,0)[lt]{\lineheight{1.25}\smash{\begin{tabular}[t]{l}a)\end{tabular}}}}%
    \put(0,0){\includegraphics[width=\unitlength,page=2]{results.pdf}}%
    \put(0.01811425,0.59747335){\makebox(0,0)[lt]{\lineheight{1.25}\smash{\begin{tabular}[t]{l}c)\end{tabular}}}}%
    \put(0,0){\includegraphics[width=\unitlength,page=3]{results.pdf}}%
    \put(0.01223406,0.31795177){\makebox(0,0)[lt]{\lineheight{1.25}\smash{\begin{tabular}[t]{l}d)\end{tabular}}}}%
    \put(0,0){\includegraphics[width=\unitlength,page=4]{results.pdf}}%
    \put(0.01811392,0.79529283){\makebox(0,0)[lt]{\lineheight{1.25}\smash{\begin{tabular}[t]{l}b)\end{tabular}}}}%
    \put(0,0){\includegraphics[width=\unitlength,page=5]{results.pdf}}%
    \put(0.01811392,0.87699499){\makebox(0,0)[lt]{\lineheight{1.25}\smash{\begin{tabular}[t]{l}b)\end{tabular}}}}%
  \end{picture}%
\endgroup%

    \vspace{-0.5cm}
\caption{\textbf{Performance Metrics Results for Visuo-Tactile Exploration during Wiping.} a) Controller shaping functions, b) Desired vs. actual motion in the base frame, c) Desired force of \SI{15}{N} shaped by the function $\rho_\mathrm{frc}$ vs. measured force in the end effector frame, d) Tank energies.}
\label{results:results}
\end{figure}

Our experimental results offer quantitative assessments across multiple performance metrics, confirming the effectiveness of our framework in exploring a surface commanded by an arbitrary tactile skill despite unknown physical constraints. We achieve a high accuracy rate in detecting the contact alignment between the robot end-effector and the surface during the exploration using the wiping task. This precision ensures reliable interaction and practical surface exploration, as depicted in Fig.~\ref{results:3d} and Fig.~\ref{results:results}.

Minimal latency ensures swift and adaptive behavior during the wiping task, enhancing overall efficiency. The vision pipeline, including pre-processing and feature extraction, achieves an average frame rate of 3 frames per second (FPS) with a loop cycle of \SI{300}{ms}. Even though \SI{300}{ms} might be considered high performance for vision processed by a standard computer, these values could be improved for better real-time perception and decision-making capabilities, essential for dynamic interaction with the environment. Additionally, the tactile perception runs at \SI{1000}{Hz}, provided by the robot's internal proprioceptive measurement. Briefly, due to the difference between the loop cycles of the two modalities, vision can be considered a spatial component in the contact monitor alignment. In contrast, the tactile sensor acts as a temporal modality. Overall, the robot finishes exploring the curvature in \SI{20}{s}. 

Accuracy in monitoring the contact alignment and exploration performance is assessed by comparing the robot's actual trajectory to the model of the contact surface, as depicted in Fig.~\ref{results:3d}. The robot is commanded only with an arbitrary force-motion policy without modeling the environment. The robot effectively explores the surface while maintaining the desired contact force. 

The robot starts without alignment or contact, as illustrated in Fig.~\ref{results:results}.a, where $\rho_\mathrm{align}$ initially decreases and then progressively increases over \SI{0}{}-\SI{5}{s}. Subsequently, it fluctuates between \SI{0}{} - \SI{1}{} whenever the contact alignment changes. $\rho_\mathrm{frc}$ decreases, particularly when the robot moves down the surface due to the margin $\delta_c$. However, the robot adjusts its posture to align with the contact when $\rho_\mathrm{align}$ is zero. Consequently, the force controller is reactivated with the updated end-effector pose, increasing $\rho_\mathrm{frc}$ to one such as at approximately around \SI{7}{s}. 

Quantitative analysis of control performance metrics reveals the absolute mean error of \SI{9}{mm}, \SI{8}{mm}, and \SI{4}{mm} in the desired wiping trajectory along the x-, y-, and z-axes, respectively (see Fig.~\ref{results:results}.b). The corresponding root mean square errors are approximately \SI{10}{mm}, \SI{10}{mm}, and \SI{6}{mm}. Further quantitative evaluation of the uniformity of pressure applied during visuo-tactile exploration is depicted in Fig.~\ref{results:results}.c, illustrating a mean absolute deviation of app. \SI{2}{N} between the desired force of \SI{15}{N} shaped by $\rho_\mathrm{frc}$ and the actual force exerted on the surface about the z-direction in the end-effector frame. Furthermore, the root mean square error in force control performance is app. \SI{3}{N}. Additionally, the contact surface has high friction, as seen from the forces about the other directions, which fluctuates up to the magnitude of \SI{10}{N}. Furthermore, Fig.~\ref{results:results}.d illustrates that the energy tanks remain within their designated limits. Specifically, $S_\mathrm{t,i}$ remains below \SI{32}{J} and decreases due to large movements on a surface with friction. Additionally, $S_\mathrm{t,f}$ remains constant at \SI{2}{J} without overloading the tank. This indicates the robot's and its surroundings' safety during contact alignment and visuo-tactile exploration.

The authors would like to mention that further study should focus on deciding $C_m$ instead of fine-tuning the current surface material properties, such as friction and rigidity. Overall, the quantitative evaluations confirm the effectiveness and practicality of our framework for automated visuo-tactile exploration of unknown rigid 3D curvatures handled at the robot's low-level control. The accuracy, precision, latency, and computational efficiency presented demonstrate our method's potential for increasing robotic deployment in manufacturing automation and other industries requiring intricate robotic interaction processes.

\section{Conclusion} \label{sec:conc}
In conclusion, this paper aims to bridge the gap between current robotic capabilities and the demands of real-world applications by simple yet effective and intuitive robotic skill programming for arbitrarily given tactile policy without requiring specialized expertise that integrates visuo-tactile exploration of unknown rigid 3D curvatures by vision-augmented unified force-impedance control. Combining tactile and vision data, we formulate a robust online contact alignment monitoring system that considers tactile error, local surface curvature, and surface orientation. This information is seamlessly integrated into a vision-augmented unified force-impedance control framework, allowing for adjusting robot stiffness and regulation of force while exploring the curvatures. Virtual energy tanks ensure system passivity and stability throughout the visuo-tactile exploration of the unknown rigid 3D curvatures.

Experimental validation with a Franka Emika research robot executing wiping tasks on challenging surfaces confirms the efficacy of our approach in achieving precise and passive visuo-tactile exploration. Comprehensive performance metrics are used for validation, including contact alignment monitoring accuracy, real-time feedback latency, computational efficiency, and control performance. These metrics provide quantitative insights into our proposed method's precision, uniformity, speed, and effectiveness, ensuring its practical applicability in real-world manufacturing scenarios. As a limitation, highly irregular curvatures are excluded from the study scope due to potential challenges for both sensors in accurately measuring surface properties. Additionally, in the current implementation, the camera can observe only the current region of interest without predicting forthcoming curvatures, a research opportunity to address in the future. Future work will investigate planning the force-motion policy for the explored curvatures toward a complete solution for generating the object-centric tactile policy for arbitrarily given policies, helping increase the robot deployment in real-world manufacturing tasks.

\section*{Acknowledgement} \label{sec:acknowledgement}
We gratefully acknowledge the funding by the European Union’s Horizon 2020 research and innovation program as part of the project ReconCycle under grant no. 871352, the Bavarian State Ministry for Economic Affairs, Regional Development and Energy (StMWi) for the Lighthouse Initiative KI.FABRIK, (Phase 1: Infrastructure and the research and development program under grant no. DIK0249), the Lighthouse Initiative Geriatronics by LongLeif GaPa gGmbH (Project Y), the German Research Foundation (DFG, Deutsche Forschungsgemeinschaft) as part of Germany’s Excellence Strategy – EXC 2050/1 – Project ID 390696704 – Cluster of Excellence “Centre for Tactile Internet with Human-in-the-Loop” (CeTI) of Technische Universität Dresden. The authors also acknowledge the financial support by the Federal Ministry of Education and Research of Germany (BMBF) in the programme of "Souverän. Digital. Vernetzt." Joint project 6G-life, project identification number 16KISK002.

\bibliographystyle{IEEEtran}
\bibliography{08_ref.bib}

\end{document}